\documentclass{article}
\usepackage{graphicx}
\PassOptionsToPackage{numbers, compress}{natbib}
\usepackage[preprint]{neurips_2019}

\usepackage[utf8]{inputenc} % allow utf-8 input
\usepackage[T1]{fontenc}    % use 8-bit T1 fonts
\usepackage{hyperref}       % hyperlinks
\usepackage{url}            % simple URL typesetting
\usepackage{booktabs}       % professional-quality tables
\usepackage{amsfonts}       % blackboard math symbols
\usepackage{nicefrac}       % compact symbols for 1/2, etc.
\usepackage{microtype}      % microtypography

\title{The Effect of Visual Design in Image Classification}

\author{
  Naftali Cohen \\
  AI Research, J.P.Morgan\\
  New York, NY 10179\\
  \texttt{naftali.cohen@jpmorgan.com} \\
    \And
  Tucker Balch \\
   AI Research, J.P.Morgan\\
  New York, NY 10179\\
  \texttt{tucker.balch@jpmorgan.com} \\
     \And
  Manuela Veloso\thanks{On leave: Machine Learning Department, Carnegie Mellon University} \\
   AI Research, J.P.Morgan\\
  New York, NY 10179\\
  \texttt{manuela.veloso@jpmorgan.com} \\ 
}

\begin{document}

\maketitle

\begin{abstract}
Financial companies continuously analyze the state of the markets to rethink and adjust their investment strategies (e.g., \cite{higgins1995analysis,de2018advances,porter2008competitive}). While the analysis is done on the digital form of data, decisions are often made based on graphical representations in white papers or presentation slides.
In this study, we examine whether binary decisions are better to be decided based on the numeric or the visual representation of the same data. 
Using two data sets, a matrix of numerical data with spatial dependencies and financial data describing the state of the S\&P index, we compare the results of supervised classification based on the original numerical representation and the visual transformation of the same data. 
We show that, for these data sets, the visual transformation results in higher predictability skill compared to the original form of the data.
We suggest thinking of the visual representation of numeric data, effectively, as a combination of dimensional reduction and feature engineering techniques (e.g., \cite{friedman2001elements}). In particular, if the visual layout encapsulates the full complexity of the data. In this view, thoughtful visual design can guard against overfitting, or introduce new features -- all of which benefit the learning process, and effectively lead to better recognition of meaningful patterns.
\end{abstract}

\section{Introduction}\label{sec:introduction}

Business companies from all sectors make daily decisions based on visual representations of data. 
In an example, financial companies may need to report their earnings, reevaluate their investment strategies, or even expose their clients to new products. What is common is that business decisions are always made after observing images that summarize the data, most commonly in the form of presentation slides or as figures in white papers %(personal communication with anonymized financial experts at a large financial institution).
(personal communication with J.P.Morgan's financial expert Nikolaos Panigirtzoglou).

Visualization of data is considered as an essential step in the data pre-processing (e.g., \cite{aggarwal2015data,bishop2006pattern,geron2017hands}). When professional data engineer is given new data to analyze, he/she will always start by visualizing the data to understand how it distributed before committing to a particular analysis strategy (e.g., \cite{wilks2011statistical,geron2017hands}). 
In this study, we aim at comparing the traditional way of analyzing data using the raw data, to that done on the visualization itself. That is, we examine the results of the same analysis when we vary the representation of the input data from raw numeric to its corresponding pixelated graphical format. 

The advantage of using raw data (e.g., \cite{aggarwal2015data,wilks2011statistical}) is that the data is kept in its objective form, uninterpolated, and not biased towards the analyst perspective or particular use case. However, for many data sets, it may be that the subjective perspective of thoughtful visual design can help to extract more insight compared to the raw data. We will show that this is, in particular, valid for data sets that hide nonlinear or interrelationships between features.

In this study, we use two different data sets. The first is synthetic numeric data, a toy example, that shows that if the hidden pattern in the data has a spatial dependency, visual representation is preferred to capture the signal in the data better. The second is real-world data set of images from a financial publication of J.P.Morgan, and we will show that their preferred way of visually representing their data has more predictability skill than its corresponding tabular form.

Previous work has shown the value in input-data transformations. 
For example, in classification problems, it is common to augment training data (i.e., resize, rotate, zoom in/out, blur, etc.) to improve the resiliency of trained models to unseen data and avoid overfitting (e.g., \cite{halevy2009unreasonable,perez2017effectiveness,krizhevsky2012imagenet,simard2003best}).
In numerical modeling, it is also common to transform data to its Fourier space to better capture the variability in the data and achieve better skill in model performances
(e.g., \cite{vallis2017atmospheric}).
In time-series classification, on the other hand, it is common to transform the data locally using wavelets and compare the various sequences according to their relevant modes of variability in the transformed space (e.g., \cite{wilks2011statistical}). 
This work is novel in that we take a direct approach in comparing the result of the same analysis applied to the various input data in their numerical and various graphical representations. 
Visual object recognition and object detection have shown great success in recent years (e.g., \cite{krizhevsky2012imagenet,koch2015siamese,lecun2015deep}) and this motivates our approach in translating a numerical problem to its graphic counterpart.

This paper is structured as follows: In section 2, we describe the data and methods, while in section 3, we show and discuss the results, and finally, we conclude the study in section 4.

\section{Data and Methods}\label{sec:data+methods}
In this study, we analyze two different data sets, synthetic data, and real-world data taken from weekly financial reports. 
We use the first data set to show an intuitive example where visualization helps in separating the binary classes. Then, following with the same reasoning, we use the real-world data to show that careful visual design of tabular numeric information can introduce feature relations that are essential for the learning.

\begin{enumerate}
\item The first data comprise of 288 samples of valid and invalid 5x5 magic squares in both their numeric matrix form and as images. 
A magic square consists of a square array of numbers consisting of the distinct positive integers $1, 2, ..., n^2$ arranged such that the sum of the $n$ numbers in any horizontal, vertical, or main diagonal line is always the same number\cite{andrews2004magic}. For a 5x5 magic squares with unique integers varying from 1 to 25, this magic constant is 65. There are 144 uniquely identified 5x5 magic squares (also called 5x5 Pan-Magic squares). However, each of the 144 unique squares has 25 translations, rotations, and reflections resulting in 28,800 magic squares\cite{rosser1939algebraic}. In this study, we focus attention on the core 144 5x5 magic squares.
In addition to the 144 5x5 magic squares, we create 144 invalid 5x5 magic squares by swapping numbers from two cells of true magic squares at random. This process results in a stratified data of valid and invalid squares. Figure 1a shows an example of a valid 5x5 magic square. 

\begin{figure}[htb]
\centering
\noindent\includegraphics[page=1,width=1\columnwidth]{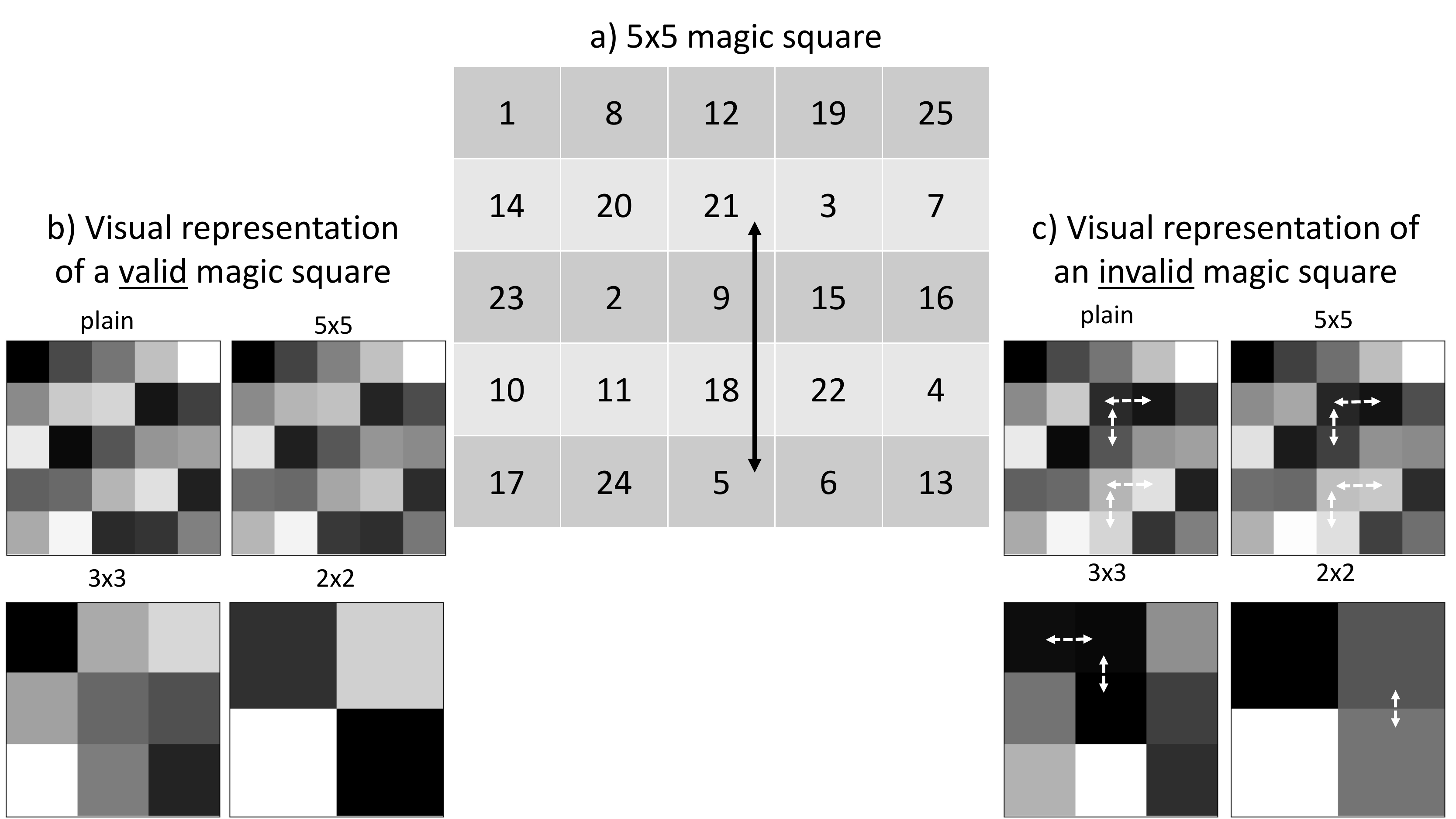}
\caption{The numerical and visual representations of a 5x5 magic square. 
}\label{fig:fig1}
\end{figure}

\item The second data comprise of market-health data which summarizes the weekly state of the S\&P 500 index\cite{de2018advances} based on five key indicators: value, positions inversed, flows, (global) economic momentum, and equity price momentum (all scaled to their corresponding z-values).
The momentum indicators take into account 2-month of variability, while flows summaries 4-weeks of retail activity. Value, of the other hand, is a long-term indicator, while positions inversed indicates for institutional investment activities. Elaborated explanations can be found in the appendix.
The data span from June 23, 2006, to April 12, 2019. Hence, this data set includes 669 rows with five features.

\begin{figure}[htb]
\centering
\noindent\includegraphics[page=2,width=1\columnwidth]{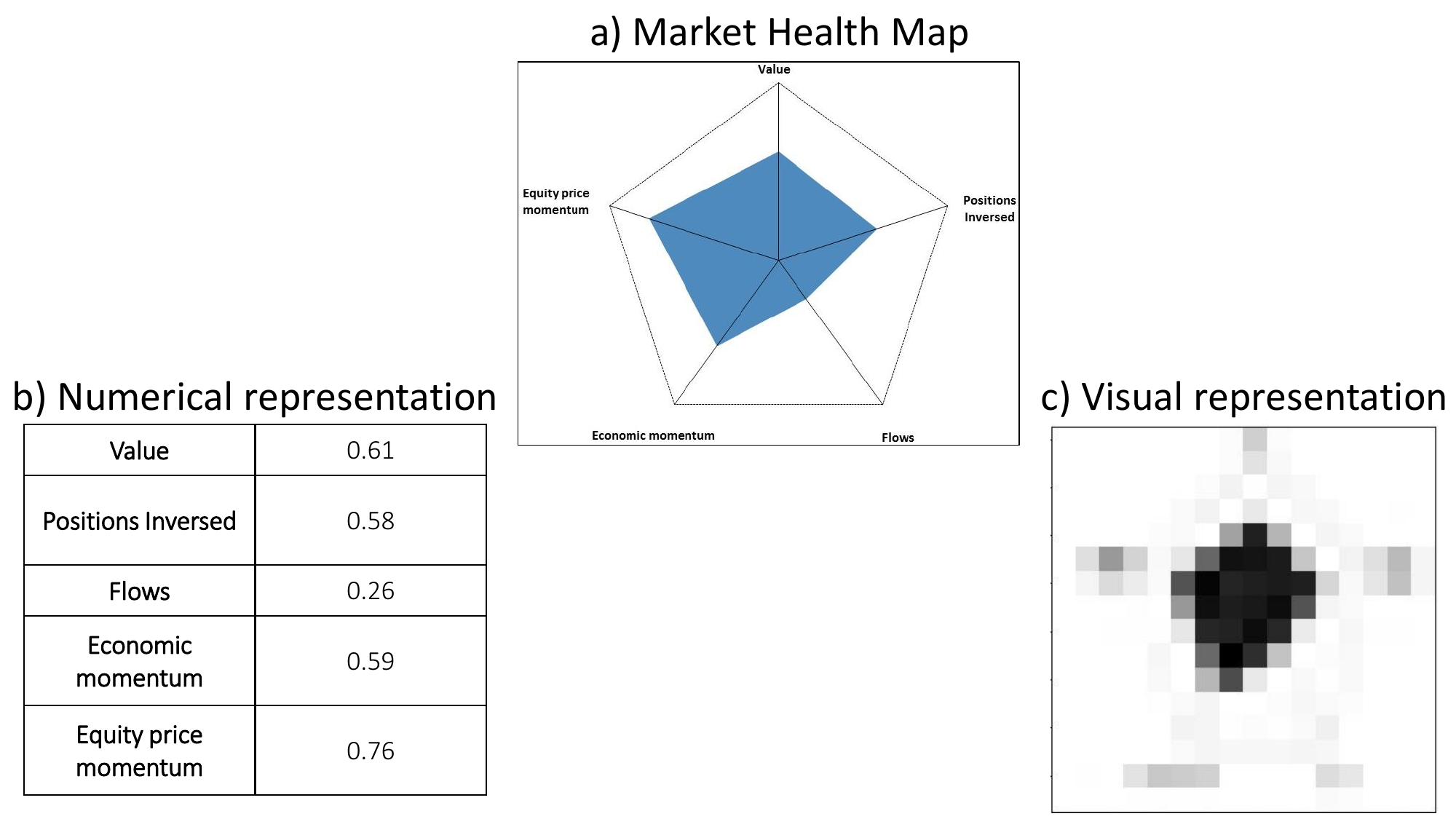}
\caption{
An example of the market-health data. 
}
\label{fig:fig2}
\end{figure}

Figure 2 shows an example of the market-health data for the week of Feb 21, 2014. 
On the left panel, one can see a table with the numerical representation of the five indicators of the market-health data. The top panel shows a filled Spider chart of the same data, and this is the way the "state of the market" is described every week in the Flows and Liquidity journal by J.P.Morgan financial experts. The image on the right is a pixelated black-and-white version of the colored image.
The data implicitly depends on time as a few indicators describe the momentum in the S\&P index, thus taking into account previous week values as well.
Publications of Flows and Liquidity can be (partially) find online\footnote{\url{https://markets.jpmorgan.com/research/email/jtkhabbm/GPS-1329637-0.pdf}}. 
This journal is mostly read by hedge funds companies, asset management firms, and individual investors.

For this data set, we will compare two label-generating rules: 
\begin{itemize}
\item Future: This labeling rule is based on the future value of the S\&P 500 index. The rule is quite simple; given next week's value of the S\&P 500 index, you decide whether to buy or sell the index this week. If the S\&P index increased in the following week, you would have wanted to buy it now, and so we label the image as Buy. On the other hand, if the S\&P 500 index decreased in the following week, we label the image as Sell.
\item Recommended: This label-generating rule is based on a logical, tree-like rule. If at least three features out of the five are above 0.5, we label the image as Buy, and otherwise as Sell.
\end {itemize}
In this way, each image or corresponding numeric row gets a binary label: buy or sell the market indicating its "health" condition.
\end{enumerate}

\section{Results and Discussion}\label{sec:results}
In this section, we will compare the result of a supervised classification prediction over both the magic-square data and the financial report images.
Both data sets are about the same size, can be represented numerically and visually, binary labeled, and, the number of samples is a least ten times larger than the number of features. The latter is a necessary condition for proper statistical inference from the data.

Let us start with the numerically represented magic-square data. This data set has 288 binary-labeled samples, each with 25 unique integers that vary from 1 to 25 (see Fig.~1a). Hence the input data has a size of 288x25. We then shuffle the samples and standardize each sample by subtracting from each value the sample mean and sample standard deviation.
We use 16 different classifiers\cite{friedman2001elements} to identify the pattern that distinguishes the two classes: Logistic Regression, Gaussian Naive-Bayes, Linear Discriminant Analysis, Quadratic Discriminant Analysis, Gaussian Process, K-Nearest Neighbors, Linear SVM, Radial Basis Function SVM, Deep Neural Net, Decision Trees, Random Forest, Extra Randomized Forest, Ada Boost, Bagging, Gradient Boosting, and Convolutional Neural Network\footnote{Using Scikit-Learn with its default parameter settings except for the Deep Neural Net which uses 32x32x32 structure, and Convolutional Neural Net (CNN) for which we use Keras with uses three layers of 32 3x3 filters with ReLU activations and Max Pooling of 2x2 in between the layers. The last layer incorporates Sigmoid activation. The CCN model is compiled with Adam optimizer, binary-cross entropy loss function and run with a batch size of 16 samples for 50 iterations}. The goal, however, is not to find the best predictive model, as a matter of fact, we spent only a little time tuning parameters for these models. The goal is to compare the aggregated performance of the models where we only change the representation of the input space. 
To evaluate the model's performance, we split and evaluate a hard voting classifier using the 10-fold cross-validation technique. This allows us to infer not only the mean prediction of the voting classifier but also the uncertainty about the prediction.

The result of the voting classifier for the numerically represented magic-squares data can be seen in Fig.~3. The mean predictability is about 0.71 in accuracy score. Note that by applying the classifiers on the flatted version of the 5x5 magic squares (288x25 input data) date, we essentially treat each cell as an independent feature and ask the classifiers to separate the 25-dimensional data into two distinct groups.
While this is not the center of this paper, it is worth noting that the best performing algorithm is the Quadratic Discriminant Analysis averaging about 0.95 in accuracy score (not shown). This is because this model assumes that the various classes have independent variability\cite{friedman2001elements}, which turns out to be an outstanding separating feature as can be seen in Table 1 (we further discuss this point later on in the text).
%\footnote{The core 5x5 magic squares have specific representation (see here:https://www.grogono.com/magic/5x5pan144.php), and so each cell's internal variability is limited. Also, the way by which we generate the invalid square (see Data and Methods Section) also limits each cell's internal variability.}. 
In any case, in this study, we are more interested in the variability of the voting classifier that aggregates various aspects in separating the classes. 

\begin{figure}[htb]
\centering
\noindent\includegraphics[page=3,width=1\columnwidth]{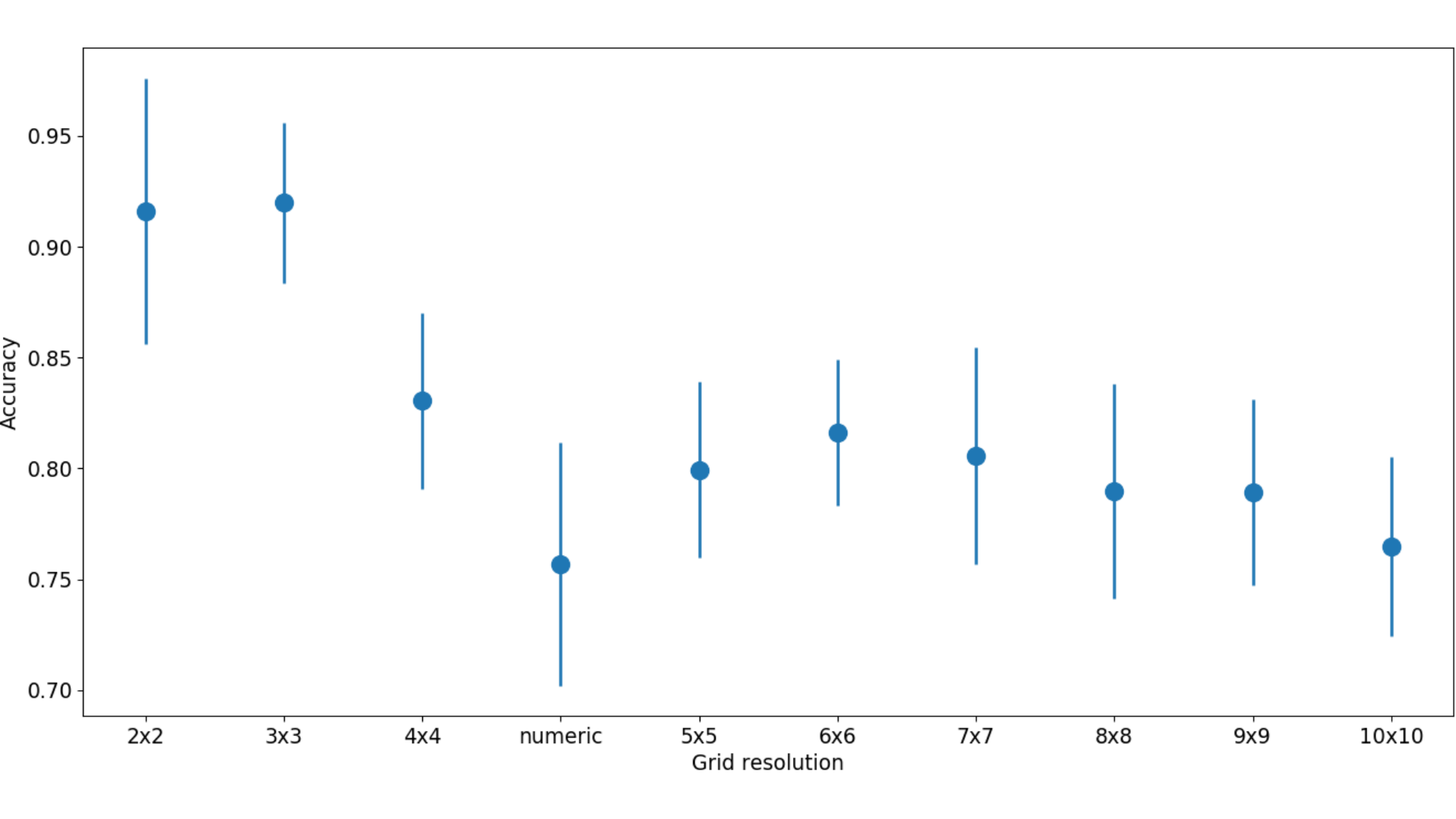}
\caption{The effect of resolution on the classification accuracy for the magic-squares data set. 
}\label{fig:fig3}
\end{figure}

The second, complementary way, to analyze the data is first to decode it, visualize, and then transform the visualization back to a digital form. While this might sound like a wasteful process, we will now show that this process serves as a combination of dimensional-reduction and feature engineering and allows to aggregate spatial information that is hidden in the data. This process also induces some random noise to the data that helps avoid overfitting.

We start this process by normalizing the numeric data (effectively, dividing each number by 25), plotting the data, and saving each sample as an image of 28x28 resolution (this specific resolution is not essential, but should be higher than 10 for our experiments, see Fig.~3). 
We then downsample each image using the Lanczos filter to resolutions that vary from 2x2 to 10x10. We are using the Lanczos, anti-aliasing filtering, as it gives smooth, high-quality down-sampling images\cite{turkowski1990filters}. 
Using the down-sampling technique, we create nine more data sets with input sizes of 288x2, 288x3, ..., 288x10. That is, we change the numeric representation of the input data to its corresponding interpolation to the various resolution grids. Note, of course, that in this particular example, the original numeric data and the 5x5 data are almost identical in their encoded information (see Fig.~1b and 1c) except for minor nuances due to the smooth interpolating nature of the Lanczos filter.

After constructing these nine new representations of the input data, we use the exact procedure as before and examine the result of the voting classifier. Figure 3 shows the result of this process. In this figure, one can see the voting-classifiers accuracy as a function of the data resolution representation, varying from low-resolution on the left to higher on the right. It is interesting to see that the highest score, about 0.9, is achieved for low-resolution input data. Besides that, the accuracy almost plateaus for resolutions higher than 5x5. There is no statistically significant difference in the accuracy of the "numeric" and the 5x5 resolution representation, as expected.

To further understand the results of Fig.~3, we would like to point the reader attention to Fig.~1b and 1c. It is easy to observe that the low-resolution invalid square data in Fig.~1c exhibit at least a pair of neighboring cells with almost no color contrast (see the white arrows). Table 1 further explain this observation by comparing the mean and standard deviation of the values in all cells of the 2x2 resolution data. It is easy to notice that both data sets have the same averaged value per cell, but the variability in the invalid magic squares is three-times larger, causing some cells to have close values. Table 1 essentially indicates that models that take into account the variability per cell can easily separate between the classes, at least on average. 

\begin{table}[htb]
\centering
\begin{tabular}{ |c | c | c|  }
 \hline
 Valid magic squares & (0,0) & (0,1)  \\
 \hline
 (0,0) & 0.50 $\pm$ 0.01  & 0.57 $\pm$ 0.01  \\
 (1,0) & 0.60 $\pm$ 0.01  & 0.47 $\pm$ 0.01  \\
 \hline
 \hline
 Invalid magic squares & (0,0) & (0,1)  \\
 \hline
 (0,0) & 0.50 $\pm$ 0.03  & 0.56 $\pm$ 0.04  \\
 (1,0) & 0.59 $\pm$ 0.03  & 0.48 $\pm$ 0.03  \\
 \hline
\end{tabular}
\caption{\label{tab:table1}The average and one standard deviation of the low-resolution 2x2 magic squares as a function of the grid location. 
}
\end{table}

By transforming the classification task into a vision problem, we achieved two things: first, the numeric data got projected to the 2-d pixel space, which is a dimensional-reduction process. This is most natural for the magic-square data as this is the way we, humans, find most natural to observe such squares (i.e., on a screen). The second is that when we downsample the pixelated data, we effectively convolve spatial information which is very informative for magic squares as their identifying pattern is spatial by summating along and across the cells -- essentially merging information from far-apart locations.
 
The analysis of the synthetic data showed us that in cases where spatial representation matters, it worth transforming the task into a vision problem to reveal hidden information.
Now, let us similarly analyze the market-health data. The market-health data is fascinating as it is numerical data that business experts found most natural to present visually in a filled circular design. The circular design is much different than a tabular form or a bar plot as it creates relationships between features. In an example, each element has now two neighboring elements, and across-variables are also proximate each other. This point is seen in Fig~2a, which reveals, in an example, the "induced" relationship between Value and Equity Price Momentum.

We repeat the same process as for the synthetic data by first standardizing the data and then fitting and validating the different classifiers, including the hard vote aggregator.
To focus the discussion, we compare the numeric form to one visual representation -- that of a 28x28 grid. For this data set, which has practical applications, we label the data in two different ways, as discussed earlier in the Data and Methods Section.

The lower row of Fig.~4 shows the accuracy scores for the multiple classifiers on the x-axes.
It can be seen that for all classifiers, there is no skill in both the visual and numerical representations when the samples are labeled using the "future" values. From that, we can argue that the weekly market-health data is not Markovian in the sense that current values do not bear enough information to deduce anything significant on what will happen next week\cite{wilks2011statistical,de2018advances}. Not even on average. This, of course, is not a surprise as we expect the market to behave close to Brownian motion, and it is quite intuitive to believe that more complex time-dependent features are required to extract the significant signal\cite{tsay2005analysis,de2018advances,hyndman2018forecasting,murphy1999technical}.

\begin{figure}[htb]
\centering
\noindent\includegraphics[page=4,width=1\columnwidth]{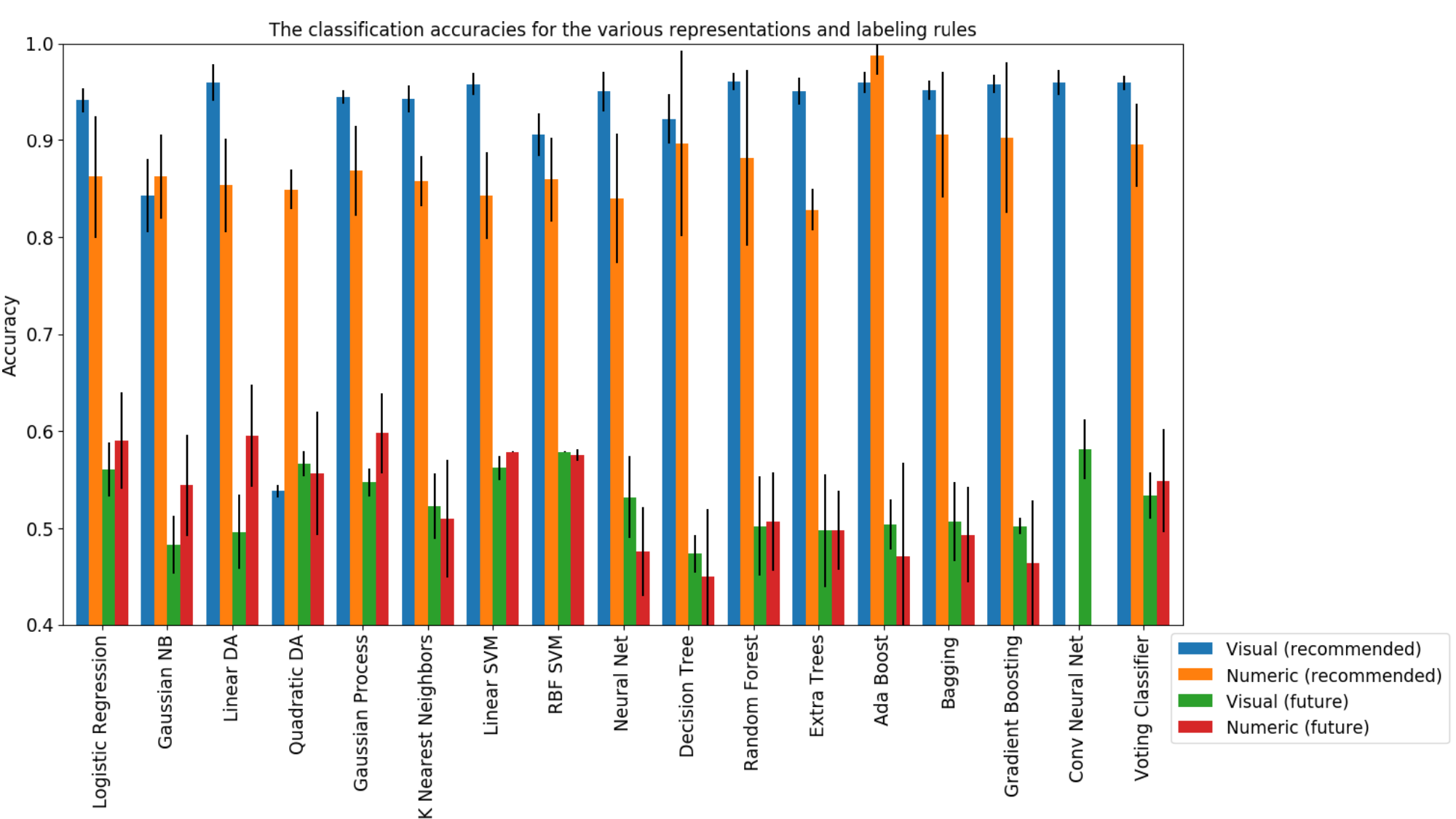}
\caption{
The accuracy scores for the different classifiers on the x-axes.
}\label{fig:fig4}
\end{figure}

While this fail in prediction was effectively time-series prediction using a one-week moving window when we label the data according to the "recommended" rule, which is a tree-like algorithm, we solve a supervised classification problem. Now Fig.~4 shows that both the visual and the numeric representations yield high accuracy scores for almost all classifiers. 
Most importantly, it shows that, on average, the visual representation does better than the corresponding numeric by about 10\%.
It worth noting that three classifiers perform better on the numeric data. The first two are the Naive Bayes and Quadratic Discriminant Analysis. Interestingly, both of these algorithms assume feature independence\cite{bishop2006pattern}. We suspect that in both of these cases the numeric tabular representation prevails because the visual representation (as seen in Fig.~2) introduces various feature dependencies (i.e., the dark pixels between proximate variables) and thus classifiers that assume feature independence, but data points have feature dependence, perform less well.
It is also worth noting that the Ada-Boost algorithm performs best on the numerical data. We suspect that this is because a tree-based algorithm with the ability to correct itself via the residual\cite{bishop2006pattern} is better on discovering the exact label-generating rule which is based on logical if conditions, thus performing better than the visual that has many more features and so noisier (28x28 compared to 25). 

\section{Conclusion}\label{sec:conclusion}
In this paper, we compared the supervised classification predication using raw tabular and visual data. 
We showed that visual representation of data allows, at times, to infer meaningful features from the specific spatial design of the visual representation.
For the magic square data set, the 2-d visual representation and the downscaling process connect far-apart information and help the learner in separating the classes.
For the market-health data, the visual design induces nonlinear feature relations that helps in separating the classes (at least for one of the label-generating rules).

We suggest thinking on visual design as, effectively, a dimensional-reduction technique. Contrary to algorithmic-based reduction methods\cite{aggarwal2015data,bishop2006pattern,wilks2011statistical,friedman2001elements} as PCA, SVD, or even feature selection -- visual representation is, in most cases, a 2-d projection of a multi-feature data based on human-intuition alone. Figure 2a is an excellent example of such a projection and its value, but, in practice, many formal business presentations include 2-d visualizations of complex multi-dimensional data.
While this is a manual process, we would like to emphasize the importance of visual design when a domain-expert design it. 
It is not as rigor, but in many aspects, we see no difference between feature selection done by a data expert to a well thought 2-d visualization done by a business professional.
The following step of choosing grid resolution amplifies the feature space to the pixel-space and creates noise\cite{aggarwal2015data}, but also nonlinear relationships and spatial dependencies.

One can argue that the decreased performance seen in Fig.~3 with increased resolution is not to do with spatial information capture, but with the fact that the ratio of the number of samples to the number of features decreased\cite{friedman2001elements}. While this effect might contribute to the decreasing trend seen in Fig.~3, we note that this ratio stays at about the same order of magnitude for all experiments. Thus this ratio change is of secondary importance. Besides, this minor change to the ratio is not expected to cause such a dramatic, statistically significant trend, as seen in Fig.~3.

\subsubsection*{Acknowledgments}
We would like to thank Nikolaos Panigirtzoglou and his team for providing us with the market-health data and for insightful comments and important ideas that helped in bringing this manuscript to completion.

\subsubsection*{Disclaimer}
Opinions and estimates constitute our judgement as of the date of this paper, are for informational purposes only and are subject to change without notice. This paper has been prepared by J.P.Morgan’s Artificial Intelligence Research Department. The goal of J.P.Morgan’s Artificial Intelligence Research Department is to explore and advance cutting-edge research in the fields of Artificial Intelligence and Machine Learning, as well as related fields like Cryptography, to develop solutions that are most impactful to J.P.Morgan’s clients and businesses. This paper is not a product of the Global Research Department of J.P.Morgan’s Corporate \& Investment Bank and therefore has not been prepared in accordance with legal requirements to disclose potential conflicts of interest or to promote the independence of investment research, including but not limited to, the prohibition on dealing ahead of the dissemination of investment research. This paper is not intended as investment research or investment advice, or a recommendation, offer or solicitation for the purchase or sale of any security, financial instrument, financial product or service, or to be used in any way for evaluating the merits of participating in any transaction. Past performance is not indicative of future results. Please consult your own advisors regarding legal, tax, accounting or any other aspects including suitability implications for your particular circumstances. J.P.Morgan disclaims any responsibility or liability whatsoever for the quality, accuracy or completeness of the information herein, and for any reliance on, or use of this material in any way.  

\bibliographystyle{plain}
\bibliography{refs}

\subsubsection*{Appendix}
Here we provide an elaborated explanation for how the five indicators in the market health map are calculated:

\begin{enumerate}

\item Positions: the difference between spec positions on US equity futures and intermediate sector UST futures. This position difference is then inverted as this signal is contrarian in nature. The higher the position in equity vs. bond futures, the lower the probability of the equity market rising in the future, and vice versa.  

\item Flow momentum: the difference between flows into equity funds (including ETFs) and flows into bond funds. We smooth the original series of reported equity minus bond fund flows using a Hodrick-Prescott filter with a lambda parameter of 100. We then take the weekly change in this smoothed series. The stronger the pace of equity vs. bond fund flows, the better the outlook for equity and risky markets.  This is because fund flows tend to be more useful as momentum signals as retail investors, the dominant investors in these funds, tend to exhibit persistence in their fund buying or selling patterns.   

\item Economic momentum: the 2-month change in the global manufacturing PMI. The global manufacturing PMI is one of the most timely and forward-looking signals for the growth trajectory of the global economy. The higher the PMI momentum, the stronger the global growth trajectory and thus, the better the outlook for equity and risky markets.  The 2-month change is more useful than the 1-month change as the latter contains a lot more noise. 

\item Equity price momentum: the 6-month change in the S\&P 500 equity index. Medium-term price momentum tends to be a useful signal for the equity market widely used by trend following investors such as CTAs.  

\item Value: the slope of the market risk-return trade off line calculated across US cash, USTs, US HG and HY corporate bonds and US equities. The slope is calculated by applying a linear regression of the Internal Rate of Return (IRR) of various assets against their historical volume. IRRs are calculated as current yield, minus expected default or downgrade losses in the case of credit. The IRR for equities is earnings yield, based on trend earnings for either operating earnings, plus the expected long-term rate of inflation. The slope of the risk-return trade-off line across major US asset classes exhibited strong and stable mean-reverting behavior over the long term and is thus more useful as a longer-term signal. Given the mean-reverting behavior of the slope of the risk-return trade offline, this signal is contrarian in nature. For example, following a long period of strong equity market performance, this market risk-return trade offline tends to flatten, i.e., the slope decreases, generating a bearish signal for the equity market. And vice versa, the risk-return trade offline tends to steepen, and its slope tends to increase following a sharp equity market correction, generating a contrarian bullish signal for the equity market.  

\end{enumerate}

\end{document}